\documentclass[letterpaper]{article} 
\usepackage{aaai2027}  
\usepackage[hyphens]{url}  
\usepackage{graphicx} 
\usepackage{natbib}  
\usepackage{caption} 

\usepackage{array}
\usepackage{multirow}
\usepackage{url}
\usepackage{pifont}
\usepackage{makecell}
\usepackage{subcaption}
\newcommand{\cmark}{\ding{51}}
\newcommand{\xmark}{\textcolor{black!45}{\ding{55}}}

\usepackage{algorithm}
\usepackage{algorithmic}

\usepackage{newfloat}
\usepackage{listings}
\DeclareCaptionStyle{ruled}{labelfont=normalfont,labelsep=colon,strut=off} 
\floatstyle{ruled}
\newfloat{listing}{tb}{lst}{}
\floatname{listing}{Listing}

\usepackage{booktabs}

\title{LabDex: A Hierarchical Benchmark for Dexterous Manipulation in Laboratories}
\author{
    Zhipeng Tang\textsuperscript{\rm 1}\equalcontrib, 
    Sihang Chen\textsuperscript{\rm 1}\equalcontrib, 
    Sha Zhang\textsuperscript{\rm 2}\corresponding, 
    Peihao Yang\textsuperscript{\rm 1}, 
    Yan Liu\textsuperscript{\rm 1}, 
    Wentao Zhao\textsuperscript{\rm 1}, 
    Xinrui Lin\textsuperscript{\rm 1}, 
    Rui Huang\textsuperscript{\rm 1}, 
    Wensheng Du\textsuperscript{\rm 1}, 
    Yuting Huang\textsuperscript{\rm 1}, 
    Jiajun Deng\textsuperscript{\rm 1}, 
    Lidian Wang\textsuperscript{\rm 1}, 
    Yuan Zhang\textsuperscript{\rm 3}, 
    Yanyong Zhang\textsuperscript{\rm 1}\corresponding
}
\affiliations{
    \textsuperscript{\rm 1}University of Science and Technology of China\\
    \textsuperscript{\rm 2}The Chinese University of Hong Kong\\
    \textsuperscript{\rm 3}iFLYTEK\\
    \{tangzhipeng, chensihang\}@mail.ustc.edu.cn, zhangsha2048@gmail.com, yanyongz@ustc.edu.cn
}

\begin{document}

\maketitle

\begingroup
\renewcommand{\thefootnote}{\fnsymbol{footnote}}
\footnotetext[0]{Dataset homepage: \url{https://zhipeng-tang.github.io/LabDex/}}
\endgroup

\begin{abstract}

Autonomous laboratories hold great promise for accelerating scientific discovery. To achieve this vision, robots are supposed to dexterously manipulate diverse labware and instruments and execute long-horizon, state-dependent experimental procedures. Yet existing benchmarks do not jointly capture dexterous hand use, real-world laboratory interactions, and multi-stage experimental procedures, limiting systematic training and evaluation. To bridge this gap, we introduce \textbf{LabDex}, a large-scale real-world dataset and benchmark for dexterous manipulation in chemistry laboratories, organized around a hierarchical task taxonomy spanning atomic skills, compositional tasks, and long-horizon experiments.
First, LabDex is cross-platform and, for the first time, unifies real-world and simulation platforms under a common framework, providing standardized task definitions, demonstrations, and evaluation protocols.
Second, LabDex is large-scale and systematically organizes chemistry laboratory operations into three interconnected levels: \emph{Atomic Skills}, which characterize fundamental dexterous manipulation capabilities; \emph{Compositional Skills}; and \emph{Long-Horizon Laboratory Workflows}.
This hierarchical design not only supports the evaluation of end-task performance, but also enables the analysis of how fundamental dexterous skills compose and influence more complex laboratory operations.
We conduct cross-level evaluations of representative robot learning methods in both real-world and simulation environments. The experimental results validate the effectiveness of the LabDex task design and demonstration data, and show that the benchmark supports the training and systematic evaluation of existing robotic policies across laboratory dexterous manipulation tasks at different levels, providing a foundation for further research and development of autonomous laboratory robots.

\end{abstract}

\begin{table*}[t]
    \centering
    \caption{Comparison with existing datasets and benchmarks.}
    \label{tab:dataset_comparison}
    \small
    \setlength{\tabcolsep}{3pt}
    \renewcommand{\arraystretch}{1.15}

    \begin{tabular}{
        @{}
        >{\raggedright\arraybackslash}p{0.25\textwidth}
        >{\centering\arraybackslash}p{0.08\textwidth}
        >{\centering\arraybackslash}p{0.08\textwidth}
        >{\centering\arraybackslash}p{0.08\textwidth}
        >{\centering\arraybackslash}p{0.08\textwidth}
        >{\centering\arraybackslash}p{0.08\textwidth}
        >{\centering\arraybackslash}p{0.12\textwidth}
        >{\centering\arraybackslash}p{0.12\textwidth}
        @{}
    }
        \toprule

        \multirow[c]{2}{*}{\textbf{Dataset / Benchmark}}
        & \multirow[c]{2}{*}{\makecell[c]{\textbf{Laboratory}\\\textbf{Domain}}}
        & \multirow[c]{2}{*}{\makecell[c]{\textbf{Dexterous}\\\textbf{Hand}}}
        & \multirow[c]{2}{*}{\textbf{Simulation}}
        & \multirow[c]{2}{*}{\textbf{Real Robot}}
        & \multicolumn{3}{c}{\textbf{Task Diversity}} \\

        \cmidrule(lr){6-8}

        & & & &
        & \textbf{Atomic}
        & \textbf{Compositional}
        & \textbf{Long-horizon} \\

        \midrule

        Bi-DexHands\cite{chen2022towards}
        & \xmark
        & \cmark
        & \cmark
        & \xmark
        & \cmark
        & \cmark
        & \xmark \\

        DexArt\cite{bao2023dexart}
        & \xmark
        & \cmark
        & \cmark
        & \xmark
        & \cmark
        & \xmark
        & \xmark \\

        DexCap\cite{wang2024dexcap}
        & \xmark
        & \cmark
        & \xmark
        & \cmark
        & \cmark
        & \cmark
        & \xmark \\

        DexVerse\cite{yao2026dexverse}
        & \xmark
        & \cmark
        & \cmark
        & \xmark
        & \cmark
        & \cmark
        & \cmark \\

        \midrule

        Chemistry3D\cite{li2024chemistry3d}
        & \cmark
        & \xmark
        & \cmark
        & \xmark
        & \cmark
        & \cmark
        & \xmark \\

        AutoBio\cite{lan2025autobio}
        & \cmark
        & \xmark
        & \cmark
        & \xmark
        & \cmark
        & \cmark
        & \xmark \\

        LabUtopia\cite{li2026labutopia}
        & \cmark
        & \xmark
        & \cmark
        & \xmark
        & \cmark
        & \cmark
        & \cmark \\

        BioProVLA-Agent\cite{du2026bioprovla}
        & \cmark
        & \xmark
        & \xmark
        & \cmark
        & \cmark
        & \cmark
        & \xmark \\

        \midrule

        \textbf{LabDex (Ours)}
        & \cmark
        & \cmark
        & \cmark
        & \cmark
        & \cmark
        & \cmark
        & \cmark \\

        \bottomrule
    \end{tabular}
\end{table*}

\section{Introduction}

Scientific laboratory automation~\cite{jiang2022artificial, abolhasani2023rise, wang2022scienceworld} requires robots to reliably interact with diverse instruments, containers, and materials.
However, the execution of complex experimental procedures remains constrained by the robot's underlying manipulation capabilities.
Compared with parallel-jaw grippers, multi-finger dexterous hands provide richer contact configurations and greater manipulation flexibility, making them a promising embodiment for laboratory operations.
Despite recent progress in dexterous manipulation, chemical laboratory scenarios remain insufficiently explored in existing datasets and benchmarks.

Chemical laboratory operations naturally exhibit a clear hierarchical structure. 
A complete experimental procedure can typically be decomposed into fundamental atomic operations with relatively independent objectives, which can be reused across different experimental tasks. Laboratory dexterous manipulation therefore involves not only the execution of individual fundamental operations, but also the ability to compose multiple operations and complete complex experimental procedures. 
Moreover, evaluating only the final outcome of a complete experimental procedure makes it difficult to distinguish a robot’s capabilities at different levels. 
Hierarchical modeling and evaluation of chemical laboratory operations can therefore provide a more fine-grained characterization of robotic dexterous manipulation capabilities across different levels of task complexity.

Recent efforts have developed robotic manipulation environments, task suites, and demonstration datasets for laboratories, while general-purpose dexterous manipulation benchmarks have expanded toward increasingly diverse and complex skills.
However, the intersection of laboratory manipulation and multi-finger dexterity remains insufficiently explored.
As shown in Table~\ref{tab:dataset_comparison}, existing benchmarks~\cite{li2024chemistry3d,lan2025autobio,li2026labutopia,du2026bioprovla} primarily employ parallel-jaw grippers or other non-dexterous robotic platforms, providing limited support for evaluating multi-finger manipulation capabilities.
Conversely, existing dexterous manipulation benchmarks~\cite{chen2022towards,bao2023dexart,wang2024dexcap,yao2026dexverse} mainly focus on generic object interactions rather than laboratory-specific operations and workflows.
Moreover, current benchmarks provide limited evaluation across atomic skills, compositional operations, and long-horizon laboratory procedures, making it difficult to characterize dexterous manipulation capabilities at different levels of task complexity.

In this work, we introduce \textbf{LabDex}, a large-scale hierarchical benchmark for dexterous manipulation in chemistry laboratories. 
Rather than treating laboratory tasks simply as independent evaluation instances, LabDex organizes laboratory manipulation hierarchically from the perspective of capability composition. 
Specifically, we divide the benchmark into three interconnected levels:
\emph{Atomic Skills}, which characterize reusable, fine-grained dexterous manipulation primitives, such as precise grasping, insertion, and liquid dispensing;
\emph{Compositional Skills}, which combine multiple atomic skills into reusable laboratory operations;
and \emph{Long-Horizon Laboratory Workflows}, which require robots to sequentially execute multiple compositional skills to complete full laboratory tasks.
Built upon unified simulation and real-world robotic platforms, LabDex provides standardized laboratory assets, task definitions, demonstration data, and evaluation protocols across all three levels.

Beyond constructing the dataset and benchmark, we also investigate how robotic dexterous manipulation capabilities scale from fundamental skills to long-horizon laboratory operations. 
Based on the proposed hierarchical evaluation framework, we systematically evaluate representative machine learning methods across all task levels. 
More importantly, we analyze the relationships among atomic skills, compositional skills, and complete laboratory workflows, revealing how failures propagate across different task levels and identifying the key capability bottlenecks that limit long-horizon laboratory manipulation. 
We envision LabDex not only as a standardized benchmark for laboratory dexterous manipulation, but also as a capability-centered evaluation framework that facilitates future research on autonomous laboratory robots.

Our contributions are summarized as follows:
\begin{itemize}
\item We introduce \textbf{LabDex}, a large-scale benchmark for laboratory dexterous manipulation that organizes operations into three hierarchical task levels: \emph{Atomic Skills}, \emph{Compositional Skills}, and \emph{Long-Horizon Laboratory Workflows}.
\item We develop a comprehensive and unified real-world and simulation evaluation platform comprising standardized laboratory assets, hierarchical task definitions, demonstration datasets, and evaluation protocols for laboratory dexterous manipulation.
\item We conduct comprehensive evaluations of representative robot learning methods and, for the first time, systematically analyze how atomic dexterous capabilities compose into long-horizon laboratory manipulation, providing new insights into the capability bottlenecks of autonomous laboratory robots.
\end{itemize}

\begin{figure*}[t]
    \centering
    \includegraphics[width=\textwidth]{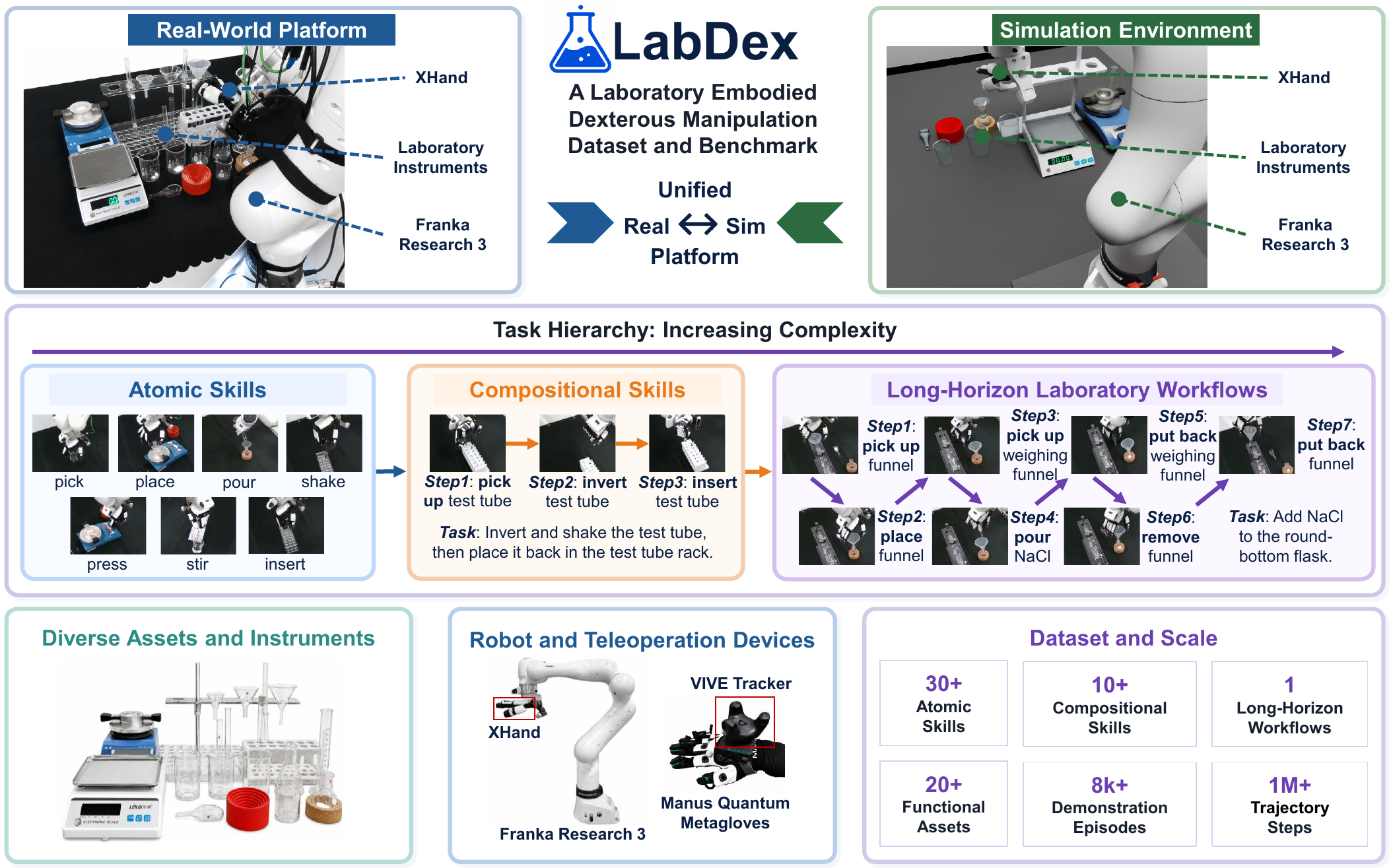}
    \caption{Overview of LabDex. LabDex is a hierarchical benchmark built on unified real-world and simulation platforms. It contains three task levels: \emph{Atomic Skills}, \emph{Compositional Skills}, and \emph{Long-Horizon Laboratory Workflows}. The dataset is collected using a Franka Research 3 robot arm equipped with an XHand dexterous hand and covers a diverse set of laboratory objects.}
    \label{fig:overview}
\end{figure*}

\section{Related Work}

\subsection{Robotic Manipulation for Laboratories}

Automated laboratories integrate machine learning, robotics, and modular platforms to improve experimental efficiency, and accelerate scientific discovery~\cite{jiang2022artificial, abolhasani2023rise}. Recent work increasingly integrates cognitive reasoning with embodied AI to enhance system autonomy and adaptivity~\cite{zhang2025position, boiko2023autonomous}. Representative systems include the Chemputer, which executes complex organic syntheses on modular hardware using a chemical programming language~\cite{steiner2019organic}; ChemCrow, which equips LLMs with expert-designed chemistry tools for planning and executing multi-step synthesis tasks~\cite{m2024augmenting}; and autonomous mobile robots that conduct exploratory synthetic chemistry using standard laboratory instruments~\cite{dai2024autonomous}. However, existing systems often depend on predefined protocols, bespoke hardware, fixed interfaces, and task-specific workstations, limiting their flexibility, generalizability, and scalability across laboratory operations. 

\subsection{Laboratory Manipulation Benchmarks}

Laboratory benchmarks require standardized tasks, reproducible settings, and comparable evaluation protocols. 
Chemistry3D provides a simulation toolkit with real-time visualization of chemical reactions~\cite{li2024chemistry3d}. 
LabUtopia combines multi-physics simulation, procedural laboratory generation, and hierarchical tasks ranging from atomic actions to long-horizon mobile manipulation~\cite{li2026labutopia}. 
AutoBio digitizes biological instruments with dedicated physics and rendering for dynamic interfaces and transparent materials, evaluating VLA policies across multiple difficulty levels~\cite{lan2025autobio}. 
However, existing laboratory manipulation benchmarks remain limited in terms of manipulation embodiment, experimental platforms, and task hierarchy.


\begin{figure*}[t]
    \centering
    \includegraphics[width=\textwidth]{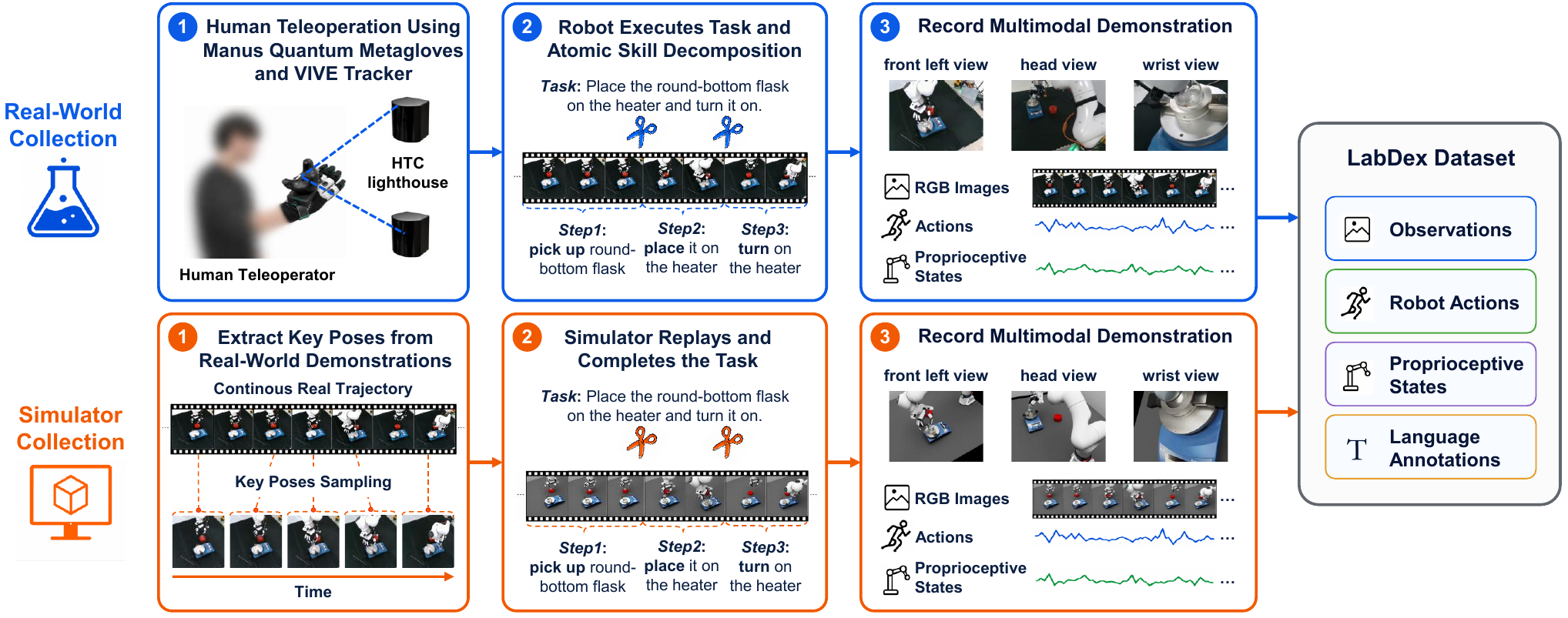}
    \caption{Real-world and simulation data collection pipeline.}
    \label{fig:pipeline}
\end{figure*}

\section{The LabDex Dataset and Benchmark}

Fig. ~\ref{fig:overview} provides an overview of LabDex. We next introduce the hierarchical task structure, unified real-world and simulation platforms, dataset statistics, and evaluation protocols.

\subsection{Hierarchical Task Structure}
LabDex tasks are derived from representative chemical laboratory procedures and are organized hierarchically according to their operational objectives and task complexity.

\paragraph{Atomic Skills.} Atomic Skills are fundamental operations with explicit objectives that can be independently evaluated and reused across tasks, including grasping, placing, insertion, pouring, stirring, and pressing. Each task has clearly defined initial states, goal states, and success criteria, and can be instantiated with different laboratory objects.

\paragraph{Compositional Skills}. Compositional Skills consist of multiple Atomic Skills executed sequentially according to the operational logic of laboratory procedures, with the goal of completing an operation with a specific experimental function. For example, the robot may grasp a container, transfer material, and return the container. This level primarily evaluates the robot’s ability to sequentially execute and compose fundamental skills.

\paragraph{Long-Horizon Laboratory Workflows}. Long-Horizon Laboratory Workflows consist of multiple Compositional Skills connected according to experimental protocols and represent complete or relatively complete experimental procedures. These tasks typically involve more laboratory objects and longer action sequences, and evaluate the robot’s ability to complete complex experimental procedures.

\subsection{Unified Real-World and Simulation Platform}

\paragraph{Real-World Hardware System.}
The LabDex real-world platform consists of a robotic execution system, visual sensing devices, and teleoperation equipment. The main hardware components are as follows:
\begin{itemize}
    \item \textbf{Franka Research 3 robot arm and XHand dexterous hand:} These serve as the primary robotic platform for executing laboratory manipulation tasks.
    \item \textbf{Intel RealSense D435i cameras:} The platform is equipped with three cameras providing wrist-view, left-front-view, and head-view observations.
    \item \textbf{Manus Quantum data glove:} The glove is used to teleoperate the XHand dexterous hand.
    \item \textbf{VIVE Tracker and Lighthouse:} These devices are used to teleoperate the Franka Research 3 robot arm.
\end{itemize}

\paragraph{Teleoperation System.}
The LabDex teleoperation system is developed based on LeFranX\cite{weng2025levr}. The robot arm is controlled using the VIVE Tracker. Specifically, the system computes the change in the tracker pose relative to its initial pose and maps this relative transformation to the target pose of the robot end effector. Inverse kinematics is then used to obtain the corresponding target joint positions of the robot arm. The dexterous hand is controlled using the Manus Quantum data glove. Human hand joint motion captured by the glove is retargeted to the target joint positions of the XHand, enabling coordinated teleoperation of the robot arm and dexterous hand.

\paragraph{Real-World Data Collection Pipeline.}
The data collection pipeline of LabDex is illustrated in Fig. \ref{fig:pipeline}. The human operator controls the robot arm and dexterous hand using the VIVE Tracker and Manus Quantum data glove, respectively. During demonstration collection, the operator segments and annotates the recorded task online according to predefined atomic-task boundaries. As a result, each complete demonstration preserves the full long-horizon laboratory workflow while also providing the corresponding compositional-skill and atomic-skill segments, resulting in a demonstration dataset with an inherent hierarchical structure.

\paragraph{Simulation Data Collection Pipeline.} As illustrated in Fig. \ref{fig:pipeline}, we first extract a set of key end-effector poses from the real-world demonstrations and subsequently select and validate them such that the robot can reproduce the corresponding tasks in simulation by executing these poses as waypoints. To support randomization of object positions and orientations, we further compute the relative transformation between each key pose and the manipulated object and represent the key poses in the object coordinate frame. When the initial pose of the manipulated object changes, the corresponding key poses can be remapped to the world coordinate frame according to the new object pose, enabling the robot to complete the task under varying object configurations. Meanwhile, the key-pose sequence is aligned with the atomic-task boundaries, allowing the generated simulation trajectories to be naturally segmented into atomic skills and thereby forming a hierarchical dataset consistent with the real-world data.

\paragraph{Demonstration Dataset.}
We use the real-world and simulation platform described above to collect expert demonstrations. Each trajectory contains the following data modalities:
\begin{itemize}
    \item \textbf{Robot proprioceptive states:} These include the joint positions, joint velocities, and end-effector pose of the robot arm, as well as the joint positions and joint torques of the dexterous hand.
    \item \textbf{Robot actions:} The robot-arm action can be represented either as $7$-dimensional target joint positions or as a $9$-dimensional end-effector pose. The latter consists of a $3$-dimensional end-effector position and a $6$-dimensional rotation representation formed by the first two rows of the rotation matrix. The dexterous-hand action includes $12$-dimensional target joint positions and $75$-dimensional human hand joint-position data. Depending on the policy formulation, either joint-space or Cartesian-space representations can be selected as the action space.
    \item \textbf{RGB images:} These consist of multi-view videos captured by three Intel RealSense D435i cameras, corresponding to the wrist, left-front, and head views. All images have a resolution of $640 \times 480$.
\end{itemize}
All modalities are synchronously recorded at $20 \mathrm{Hz}$.

\subsection{Dataset Tasks and Statistics}
LabDex consists of $7$ atomic-skill categories, instantiated into over $30$ atomic skills across over $20$ laboratory objects. These atomic tasks are further organized into over $10$ compositional skills and $1$ long-horizon laboratory workflows. The benchmark covers both simulation and real-world settings, with standardized task definitions across the three capability levels. More details can be found in Appendix C.

\begin{table}[t]
    \centering
    \caption{Average performance of atomic skill categories. We evaluate a total of 26 atomic skill instances, with each task evaluated over 50 trials for every model. Complete results for all atomic skills are provided in Appendix B.}
    \label{tab:atomic_skill_result}
    \small
    \setlength{\tabcolsep}{4pt}
    \renewcommand{\arraystretch}{1.15}

    \begin{tabular*}{\columnwidth}{
        @{\extracolsep{\fill}}
        lccc
        @{}
    }
        \toprule
        \textbf{Skill Type}
        & \textbf{DP}
        & \textbf{ACT}
        & ${\pi}_{0.5}$ \\
        \midrule

        Pick
        & $0.05$ & $0.25$ & $\mathbf{0.46}$ \\

        Place
        & $0.01$ & $0.64$ & $\mathbf{0.92}$ \\

        Insert
        & $0.00$ & $0.25$ & $\mathbf{0.30}$ \\

        Pour
        & $0.00$ & $0.23$ & $\mathbf{0.66}$ \\

        Press
        & $0.00$ & $0.00$ & $\mathbf{0.12}$ \\

        Shake
        & $0.00$ & $\mathbf{0.90}$ & $0.88$ \\

        Stir
        & $0.00$ & $0.00$ & $\mathbf{0.46}$ \\

        \midrule

        \textbf{Overall Average}
        & $0.02$ & $0.34$ & $\mathbf{0.57}$ \\

        \bottomrule
    \end{tabular*}
\end{table}

\begin{table*}[!t]
    \centering
    \caption{Performance of compositional skills. Each task is evaluated over 50 trials for every model.}
    \label{tab:compositional_skill_result}
    \small
    \setlength{\tabcolsep}{4pt}
    \renewcommand{\arraystretch}{1.15}
    
    \begin{tabular*}{\textwidth}{
        @{\extracolsep{\fill}}
        >{\raggedright\arraybackslash}p{0.18\textwidth}
        lccc
        @{}
    }
        \toprule
        \textbf{Task}
        & \textbf{Sub-task}
        & \textbf{DP}
        & \textbf{ACT}
        & ${\pi}_{0.5}$ \\
        \midrule






        \multirow[c]{4}{0.18\textwidth}{%
            \makecell[l]{
                pour\_the\_water\_from\\
                the\_right\_beaker\_into\\
                the\_left\_one
            }%
        }
        & pick\_up\_the\_beaker\_on\_the\_right
        & $0.00$ & $0.70$ & $0.62$ \\

        & pour\_the\_water\_from\_the\_handheld\_beaker\_into\_the\_one\_on\_the\_table
        & $0.00$ & $0.48$ & $0.32$ \\

        & place\_the\_beaker\_back\_on\_the\_table
        & $0.00$ & $0.48$ & $0.00$ \\

        & \textbf{Avg. Len.}
        & $0.00$ & $\mathbf{1.66}$ & $0.94$ \\

        \midrule

        \multirow[c]{3}{0.18\textwidth}{%
            \makecell[l]{
                put\_the\_100ml\_beaker\\
                on\_the\_electronic\\
                balance
            }%
        }
        & pick\_up\_the\_100ml\_beaker
        & $0.00$ & $0.12$ & $0.72$ \\

        & put\_the\_100ml\_beaker\_on\_the\_electronic\_balance
        & $0.00$ & $0.12$ & $0.70$ \\

        & \textbf{Avg. Len.}
        & $0.00$ & $0.24$ & $\mathbf{1.42}$ \\

        \midrule

        \multirow[c]{3}{0.18\textwidth}{%
            \makecell[l]{
                put\_the\_200ml\_beaker\\
                on\_the\_electronic\\
                balance
            }%
        }
        & pick\_up\_the\_200ml\_beaker
        & $0.00$ & $0.62$ & $0.90$ \\

        & put\_the\_200ml\_beaker\_on\_the\_electronic\_balance
        & $0.00$ & $0.54$ & $0.88$ \\

        & \textbf{Avg. Len.}
        & $0.00$ & $1.16$ & $\mathbf{1.78}$ \\

        \midrule

        \multirow[c]{4}{0.18\textwidth}{%
            \makecell[l]{
                shake\_the\_test\_tube\\
                and\_put\_it\_back\\
                in\_the\_rack
            }%
        }
        & pick\_up\_the\_test\_tube
        & $0.00$ & $0.12$ & $0.14$ \\

        & shake\_the\_test\_tube
        & $0.00$ & $0.02$ & $0.14$ \\

        & put\_the\_test\_tube\_back\_in\_the\_rack
        & $0.00$ & $0.00$ & $0.02$ \\

        & \textbf{Avg. Len.}
        & $0.00$ & $0.14$ & $\mathbf{0.30}$ \\

        \midrule

        \multirow[c]{4}{0.18\textwidth}{%
            \makecell[l]{
                stir\_the\_solution\\
                in\_the\_beaker\_with\\
                a\_glass\_rod
            }%
        }
        & take\_the\_glass\_rod\_from\_the\_white\_test\_tube\_rack
        & $0.00$ & $0.26$ & $0.14$ \\

        & stir\_in\_the\_beaker\_to\_the\_left\_of\_the\_white\_test\_tube\_rack\_with\_the\_glass\_rod
        & $0.00$ & $0.08$ & $0.08$ \\

        & put\_the\_glass\_rod\_back\_on\_the\_white\_test\_tube\_rack
        & $0.00$ & $0.00$ & $0.02$ \\

        & \textbf{Avg. Len.}
        & $0.00$ & $\mathbf{0.34}$ & $0.24$ \\

        \midrule

        \multirow[c]{4}{0.18\textwidth}{%
            \makecell[l]{
                pour\_the\_water\_from\\
                the\_graduated\_cylinder\\
                into\_the\_round\_bottom\\
                flask
            }%
        }
        & pick\_up\_the\_graduated\_cylinder
        & $0.00$ & $0.00$ & $0.00$ \\

        & pour\_the\_water\_from\_the\_graduated\_cylinder\_into\_the\_round\_bottom\_flask
        & $0.00$ & $0.00$ & $0.00$ \\

        & put\_the\_graduated\_cylinder\_back\_on\_the\_table
        & $0.00$ & $0.00$ & $0.00$ \\

        & \textbf{Avg. Len.}
        & $0.00$ & $0.00$ & $0.00$ \\

        \midrule

        \multirow[c]{3}{0.18\textwidth}{%
            \makecell[l]{
                place\_the\_funnel\_on\\
                the\_round\_bottom\_flask
            }%
        }
        & pick\_up\_the\_funnel\_from\_the\_transparent\_funnel\_stand
        & $0.00$ & $0.60$ & $0.94$ \\

        & place\_the\_funnel\_on\_the\_round\_bottom\_flask
        & $0.00$ & $0.00$ & $0.10$ \\

        & \textbf{Avg. Len.}
        & $0.00$ & $0.60$ & $\mathbf{1.04}$ \\

        \midrule

        \multirow[c]{4}{0.18\textwidth}{%
            \makecell[l]{
                pour\_the\_NaCl\_from\\
                the\_weighing\_funnel\\
                into\_the\_round\_bottom\\
                flask
            }%
        }
        & pick\_up\_the\_weighing\_funnel
        & $0.00$ & $0.12$ & $0.14$ \\

        & pour\_all\_the\_NaCl\_from\_the\_weighing\_funnel\_into\_the\_round\_bottom\_flask
        & $0.00$ & $0.00$ & $0.02$ \\

        & put\_the\_weighing\_funnel\_back\_on\_the\_table
        & $0.00$ & $0.00$ & $0.02$ \\

        & \textbf{Avg. Len.}
        & $0.00$ & $0.12$ & $\mathbf{0.18}$ \\

        \midrule

        \multirow[c]{3}{0.18\textwidth}{%
            \makecell[l]{
                put\_the\_funnel\_back\\
                on\_the\_funnel\_stand
            }%
        }
        & remove\_the\_funnel\_from\_the\_round\_bottom\_flask
        & $0.00$ & $0.28$ & $0.46$ \\

        & put\_the\_funnel\_back\_on\_the\_funnel\_stand
        & $0.00$ & $0.10$ & $0.24$ \\

        & \textbf{Avg. Len.}
        & $0.00$ & $0.38$ & $\mathbf{0.70}$ \\

        \midrule
        
        \multicolumn{2}{l}{\textbf{Overall Avg. Len.}}
        & $0.00$ & $0.52$ & $\mathbf{0.73}$ \\

        \bottomrule
    \end{tabular*}
\end{table*}

\subsection{Evaluation Protocol}
The positions of the manipulated objects are randomly initialized within a region determined by the spatial range covered during training data collection. A task instance is considered successful if, after the final task stage is completed, the current state remains within the tolerance threshold of the target state for two consecutive seconds. This success criterion is consistent with prior work~\cite{li2026labutopia, gong2023arnold, he2024learning, hu2024video, liu2024robomamba}. 

For \emph{Atomic Skills}, LabDex uses task success rate as the evaluation metric. For \emph{Compositional Skills} and \emph{Long-Horizon Laboratory Workflows}, in addition to reporting the overall success rate of the complete task, we measure the success rate of each atomic skill and compute the average number of atomic skills completed per evaluation rollout.

\section{Experiments}

\subsection{Experimental Setup}
\paragraph{Models.}
To benchmark the performance of existing algorithms in LabDex, we select three representative models: ACT~\cite{zhao2023learning}, Diffusion Policy~\cite{chi2025diffusion} and $\pi_{0.5}$~\cite{intelligence2025pi05}.

\paragraph{Training Details.} For DP and ACT, we directly adopt the implementations provided in the LeRobot\cite{cadene2026lerobot} repository, while $\pi_{0.5}$ is implemented using the official OpenPI codebase. All model training and deployment are conducted on NVIDIA RTX 3090 GPUs.
Specifically, DP and ACT are trained on a single GPU with a batch size of 16. The number of training steps is determined according to the task level: 100,000 steps for \emph{Atomic Skills}, 200,000 steps for \emph{Compositional Skills}, and 400,000 steps for \emph{Long-Horizon Laboratory Workflows}. For $\pi_{0.5}$, we use two GPUs with a total batch size of 16. The model is trained for 30,000 steps on \emph{Atomic Skills}, 60,000 steps on \emph{Compositional Skills}, and 100,000 steps on \emph{Long-Horizon Laboratory Workflows}.
In addition, ACT and $\pi_{0.5}$ condition their predictions on a single observation frame, whereas DP uses two consecutive observation frames as input. The action chunk sizes of DP, ACT, and $\pi_{0.5}$ are set to $48$, $48$, and $50$, respectively. All models use joint angles as actions. All tasks are trained using 200 demonstrations by default.

\paragraph{Evaluation Details.} Each task is evaluated over 50 trials for every model. The positions of the manipulated objects and the robot arm poses are randomized within the ranges covered by the training data.

\noindent
More details can be found in Appendix A. 

\subsection{Experimental Results}

\paragraph{Results of Atomic Skills.} The experimental results for \emph{Atomic Skills} are presented in Tab. \ref{tab:atomic_skill_result}. Complete results for all atomic skills are provided in Appendix B. Overall, $\pi_{0.5}$ achieves the highest average success rate of $0.57$, substantially outperforming ACT at $0.34$ and DP at $0.02$. $\pi_{0.5}$ performs best on most tasks and operation categories, particularly on placement tasks, where it achieves an average success rate of $0.92$. In contrast, ACT is more sensitive to the initial object positions during evaluation: it may achieve high success rates in certain regions of the workspace, while its performance degrades considerably in others. This localized performance variation can also allow ACT to outperform $\pi_{0.5}$ on a small number of challenging tasks.

In terms of operation type, placement tasks are generally easier. These tasks typically require the robot only to move an already grasped object to a target region, whereas grasping and insertion tasks impose greater demands on finger configurations and object-contact relationships, resulting in lower success rates. Pressing tasks achieve the lowest success rate, which may be attributed to the lack of tactile sensing.

In terms of manipulated objects, tasks involving glass rods, test tubes, and graduated cylinders, which have small grasping regions or elongated geometries, are generally more challenging. These objects are more sensitive to grasp locations, finger-closing patterns, and end-effector pose errors, and therefore requiring greater manipulation accuracy.

\paragraph{Results of Compositional Skills.} The experimental results for \emph{Compositional Skills} are presented in Tab. \ref{tab:compositional_skill_result}. Overall, $\pi_{0.5}$ achieves the highest average number of completed atomic skills, reaching $0.73$, compared with $0.52$ for ACT, while DP fails to complete any atomic skill across all compositional tasks. The results reveal two main bottlenecks that hinder the completion of compositional skills. First, some tasks are constrained by the initial grasping of objects such as round-bottom flasks, graduated cylinders, test tubes, and glass rods. Second, even when the robot successfully completes the initial grasp, its performance often declines substantially during subsequent skill transitions, particularly for operations such as insertion and returning an object after pouring. For example, $\pi_{0.5}$ achieves a relatively high success rate when grasping the funnel, but its success rate drops considerably when subsequently inserting the funnel into the round-bottom flask. These results indicate that the key limitations of existing methods lie not only in mastering individual atomic skills, but also in maintaining stable object states and reliably composing multiple skills into continuous execution.

\paragraph{Results of Long-Horizon Laboratory Workflows.} The results for \emph{Long-Horizon Laboratory Workflows} are presented in Tab. \ref{tab:long_horizon_workflow_result}. Overall, $\pi_{0.5}$ achieves the highest average number of completed atomic skills, reaching $0.62$, compared with $0.40$ for ACT, while DP fails to complete any atomic skill. However, none of the three models is able to complete the full workflow, indicating that existing methods still face substantial challenges in long-horizon laboratory manipulation. Further analysis shows that the primary bottleneck occurs during the insertion stage following the initial funnel grasp. Although ACT and $\pi_{0.5}$ can complete the initial grasp with some success, their performance drops substantially when placing the funnel onto the round-bottom flask, preventing the subsequent operations from being executed.

\paragraph{Results in Simulation.} In the simulation experiments, we collect 50 demonstrations for each task and use them to train each model separately. Partial results for Compositional Skills are presented in Tab. \ref{tab:sim_compositional_task_result}, showing an overall trend consistent with that observed in the real-world experiments, with $\pi_{0.5}$ achieving the best performance. For tasks involving grasping and pouring with a graduated cylinder, failures in the initial grasp constitute the primary bottleneck, preventing the execution of subsequent operations. This observation is consistent with the findings from the real-world experiments.

\noindent 
More experimental results can be found in Appendix B.

\begin{table*}[!t]
    \centering
    \caption{Performance of long-horizon laboratory workflows. Each task is evaluated over 50 trials for every model.}
    \label{tab:long_horizon_workflow_result}
    \small
    \setlength{\tabcolsep}{4pt}
    \renewcommand{\arraystretch}{1.15}
    
    \begin{tabular*}{\textwidth}{
        @{\extracolsep{\fill}}
        >{\raggedright\arraybackslash}p{0.18\textwidth}
        lccc
        @{}
    }
        \toprule
        \textbf{Task}
        & \textbf{Sub-task}
        & \textbf{DP}
        & \textbf{ACT}
        & ${\pi}_{0.5}$ \\
        \midrule

        \multirow[c]{8}{0.18\textwidth}{%
            \makecell[l]{
                add\_the\_NaCl\_from\\
                the\_weighing\_funnel\\
                to\_the\_round\_bottom\\
                flask
            }%
        }
        & pick\_up\_the\_funnel\_from\_the\_transparent\_funnel\_stand
        & $0.00$ & $0.40$ & $0.54$ \\

        & place\_the\_funnel\_on\_the\_round\_bottom\_flask
        & $0.00$ & $0.00$ & $0.02$ \\

        & pick\_up\_the\_weighing\_funnel
        & $0.00$ & $0.00$ & $0.02$ \\

        & pour\_all\_the\_NaCl\_from\_the\_weighing\_funnel\_into\_the\_round\_bottom\_flask
        & $0.00$ & $0.00$ & $0.02$ \\

        & put\_the\_weighing\_funnel\_back\_on\_the\_table
        & $0.00$ & $0.00$ & $0.02$ \\

        & remove\_the\_funnel\_from\_the\_round\_bottom\_flask
        & $0.00$ & $0.00$ & $0.00$ \\

        & put\_the\_funnel\_back\_on\_the\_funnel\_stand
        & $0.00$ & $0.00$ & $0.00$ \\

        & \textbf{Avg. Len.}
        & $0.00$ & $0.40$ & $\mathbf{0.62}$ \\

        \bottomrule
    \end{tabular*}
\end{table*}

\begin{table*}[!t]
    \centering
    \caption{Performance of compositional skills in simulation. Each task is evaluated over 50 trials for every model.}
    \label{tab:sim_compositional_task_result}
    \small
    \setlength{\tabcolsep}{4pt}
    \renewcommand{\arraystretch}{1.15}

    \begin{tabular*}{\textwidth}{
        @{\extracolsep{\fill}}
        >{\raggedright\arraybackslash}p{0.18\textwidth}
        lccc
        @{}
    }
        \toprule
        \textbf{Task}
        & \textbf{Sub-task}
        & \textbf{DP}
        & \textbf{ACT}
        & ${\pi}_{0.5}$ \\
        \midrule

        \multirow[c]{3}{0.18\textwidth}{%
            \makecell[l]{
                put\_the\_200ml\_beaker\\
                on\_the\_electronic\\
                balance
            }%
        }
        & pick\_up\_the\_200ml\_beaker
        & $0.00$ & $0.26$ & $0.58$ \\

        & put\_the\_200ml\_beaker\_on\_the\_electronic\_balance
        & $0.00$ & $0.24$ & $0.54$ \\

        & \textbf{Avg. Len.}
        & $0.00$ & $0.46$ & $\mathbf{1.12}$ \\

        \midrule

        \multirow[c]{4}{0.18\textwidth}{%
            \makecell[l]{
                pour\_the\_water\_from\\
                the\_graduated\_cylinder\\
                into\_the\_round\_bottom\\
                flask
            }%
        }
        & pick\_up\_the\_graduated\_cylinder
        & $0.00$ & $0.00$ & $0.00$ \\

        & pour\_the\_water\_from\_the\_graduated\_cylinder\_into\_the\_round\_bottom\_flask
        & $0.00$ & $0.00$ & $0.00$ \\

        & put\_the\_graduated\_cylinder\_back\_on\_the\_table
        & $0.00$ & $0.00$ & $0.00$ \\

        & \textbf{Avg. Len.}
        & $0.00$ & $0.00$ & $0.00$ \\

        \bottomrule
    \end{tabular*}
\end{table*}

\begin{table*}[!t]
    \centering

    \begin{minipage}[t]{0.22\textwidth}
        \vspace{0pt}
        \centering
        \captionof{table}{Performance under Different Numbers of Demonstrations.}
        \label{tab:data_skill_result}
        \renewcommand{\arraystretch}{1.8}
        \setlength{\tabcolsep}{4pt}

        \resizebox{\linewidth}{!}{%
            \begin{tabular}{lccc}
                \toprule
                \textbf{Episodes}
                & \textbf{DP}
                & \textbf{ACT}
                & ${\pi}_{0.5}$ \\
                \midrule
                $50$  & $0.00$ & $0.00$  & $0.30$  \\
                $100$ & $0.00$ & $0.14$ & $0.34$  \\
                $200$ & $0.00$ & $0.36$ & $1.00$ \\
                \bottomrule
            \end{tabular}
        }
    \end{minipage}
    \hfill
    \begin{minipage}[t]{0.40\textwidth}
        \vspace{0pt}
        \centering
        \footnotesize
        \captionof{table}{Performance under object distractions.}
        \label{tab:object_distraction_result}
        \renewcommand{\arraystretch}{1.8}
        \setlength{\tabcolsep}{4pt}

        \begin{tabular}{
            >{\raggedright\arraybackslash}p{0.72\linewidth}
            >{\centering\arraybackslash}p{0.20\linewidth}
        }
            \toprule
            \textbf{Distractor Objects}
            & \textbf{Succ. Rate} \\
            \midrule
            None
            & $1.00$ \\

            100ml beaker
            & $0.00$ \\

            Graduated cylinder
            & $0.42$ \\

            Graduated cylinder and conical flask
            & $0.16$ \\
            \bottomrule
        \end{tabular}
    \end{minipage}
    \hfill
    \begin{minipage}[t]{0.30\textwidth}
        \vspace{0pt}
        \centering
        \footnotesize
        \captionof{table}{Performance on novel objects.}
        \label{tab:novel_object_generalization_result}
        \renewcommand{\arraystretch}{1.8}
        \setlength{\tabcolsep}{4pt}

        \resizebox{\linewidth}{!}{%
            \begin{tabular}{lcc}
                \toprule
                \textbf{Object}
                & \textbf{Novel Object}
                & \textbf{Succ. Rate} \\
                \midrule
                200ml beaker
                & \xmark & $0.88$ \\

                100ml beaker
                & \cmark & $0.96$ \\

                Reagent bottle
                & \cmark & $0.20$ \\

                Weighing funnel
                & \cmark & $0.22$ \\

                Three-neck flask
                & \cmark & $0.00$ \\
                \bottomrule
            \end{tabular}
        }
    \end{minipage}
\end{table*}

\subsection{Ablation Study}
\paragraph{Training Data Scale.} The results under different amounts of training data are presented in Tab. \ref{tab:data_skill_result}. As the number of demonstrations increases from 50 to 200, the performance of both ACT and $\pi_{0.5}$ improves. In particular, $\pi_{0.5}$ achieves a success rate of $1.00$ with 200 demonstrations, while ACT improves from $0$ to $0.36$. These results indicate that increasing the amount of demonstration data can effectively improve task learning, and that stronger models benefit more substantially from larger-scale training data.

\paragraph{Object Distractors.} The object-distractor experiment is conducted on the pick\_up\_the\_200ml\_beaker task using $\pi_{0.5}$, with the results shown in Tab. \ref{tab:object_distraction_result}. Without distractor objects, the model achieves a success rate of 1.00. Its performance decreases substantially after distractors are introduced. When a graduated cylinder is added to the scene, the success rate drops to 0.42, and further decreases to 0.16 when an erlenmeyer flask is also included. In particular, adding only a 100 ml beaker, which is highly similar to the target object, reduces the success rate to $0$. This suggests that visually and geometrically similar objects cause stronger interference with target recognition and selection.

\paragraph{Novel-Object Generalization.}
We jointly train $\pi_{0.5}$ on six grasping tasks and evaluate its generalization to novel objects. As shown in Tab.~\ref{tab:novel_object_generalization_result}, it achieves a success rate of $0.96$ on an unseen 100 ml beaker, demonstrating effective generalization to objects with similar shapes and manipulation requirements. However, performance drops substantially as geometric differences from the training objects increase, indicating that its generalization remains largely limited to similar objects.

\section{Conclusion}

We introduce \textbf{LabDex}, a hierarchical benchmark for dexterous manipulation in chemistry laboratories. LabDex is built on unified real-world and simulation platforms and organizes laboratory manipulation tasks into three interconnected levels: \emph{Atomic Skills}, \emph{Compositional Skills}, and \emph{Long-Horizon Laboratory Workflows}. 
We evaluate representative robot learning methods on LabDex. The experimental results demonstrate that reliable fundamental grasping, stable skill composition, and stronger generalization capabilities remain key challenges in realizing autonomous laboratory robots. We hope that LabDex will provide a standardized foundation for training, evaluating, and comparing algorithms for laboratory dexterous manipulation, while facilitating the development of more reliable and generalizable autonomous laboratory robotic systems.

\bibliography{aaai2027}

\clearpage
\appendix

\twocolumn[
\begin{center}
    {\LARGE\bfseries Appendix}
\end{center}
\vspace{1em}
]

\section{Implementation Details}

\paragraph{Additional Details about Hardware Systems.} The LabDex real-world platform consists of a robotic execution system, visual sensing devices, teleoperation equipment and two workstations. The main hardware components are as follows:
\begin{itemize}
    \item \textbf{Franka Research 3 robot arm and XHand dexterous hand:} These serve as the primary robotic platform for executing laboratory manipulation tasks.
    \item \textbf{Intel RealSense D435i cameras:} The platform is equipped with three cameras providing wrist-view, left-front-view, and head-view observations.
    \item \textbf{Manus Quantum data glove:} The glove is used to teleoperate the XHand dexterous hand.
    \item \textbf{VIVE Tracker and Lighthouse:} These devices are used to teleoperate the Franka Research 3 robot arm.
    \item \textbf{Ubuntu workstation:} This workstation serves as the main control and data collection host, and is responsible for recording experimental data, controlling the robot arm and dexterous hand, and deploying robotic policies.
    \item \textbf{Windows workstation:} This workstation connects to the Manus Quantum data glove, receives the captured human hand motion data, and transmits the data to the Ubuntu workstation.
\end{itemize}

\paragraph{Additional Details about Baseline Models.} We select three representative models: ACT, Diffusion Policy and $\pi_{0.5}$.
\begin{itemize}
    \item \textbf{ACT.} ACT integrates multi-view visual observations with the robot's current joint states through a transformer architecture and outputs a fixed-length chunk of future target joint positions. During execution, predictions from overlapping chunks are aggregated through temporal ensembling, enabling smoother control while alleviating error accumulation over long task horizons.

    \item \textbf{Diffusion Policy.} Diffusion Policy models robot control as a conditional denoising diffusion process. Given recent observations, it iteratively refines a noisy action sequence into a future control trajectory.
    
    \item \textbf{$\pi_{0.5}$.} $\pi_{0.5}$ is a vision-language-action model designed for open-world robotic manipulation. Given visual observations, language instructions, and robot states, it first predicts a high-level subtask and then generates a low-level action chunk using a flow-matching action expert.
\end{itemize}

\paragraph{Evaluation Details.} Each task is evaluated over 50 trials for every model. For Atomic Skills, we consider two initialization settings according to their positions in the complete task sequence:
\begin{itemize}
    \item If the evaluated atomic skill is the first atomic operation in the task sequence, the robot arm and dexterous hand are reset to their default initial states, while only the positions of the manipulated objects involved in the skill are randomized.
    \item If the evaluated atomic skill is not the first atomic operation in the task sequence, all preceding atomic operations are first replayed to bring the environment to the required initial state. The end-effector pose is then randomized within the range covered by the training data.
\end{itemize}
Compositional Skills are evaluated using the same initialization protocol as Atomic Skills. For Long-Horizon Laboratory Workflows, only the positions of all manipulated objects involved in the task are randomized.

\section{Additional Experimental Results}

\paragraph{Complete Real-World Atomic Skills Results.} As shown in Tab. \ref{tab:atomic_skill_result}, we report the complete results for Atomic Skills in the real-world setting. Each task is evaluated over 50 trials for every model.

\paragraph{Simulation Atomic Skills Results.} As shown in Tab. \ref{tab:sim_atomic_skill_result}, we report the results for Atomic Skills in the simulation setting. Each task is evaluated over 50 trials for every model.

\begin{table*}[!t]
    \centering
    \caption{Performance of atomic skills. Each task is evaluated over 50 trials for every model.}
    \label{tab:atomic_skill_result}
    \small
    \setlength{\tabcolsep}{3pt}
    \renewcommand{\arraystretch}{1.15}

    \begin{tabular*}{\textwidth}{
        @{\extracolsep{\fill}}
        llccc
        @{}
    }
        \toprule
        \textbf{Type}
        & \textbf{Task}
        & \textbf{DP}
        & \textbf{ACT}
        & ${\pi}_{0.5}$ \\
        \midrule
    
        \multirow{11}{*}{Pick}
        & pick\_up\_the\_round\_bottom\_flask
        & $0.00$ & $0.02$ & $0.12$ \\
    
        & pick\_up\_the\_beaker\_on\_the\_right
        & $0.38$ & $0.38$ & $0.86$ \\
    
        & pick\_up\_the\_100ml\_beaker
        & $0.00$ & $0.02$ & $0.82$ \\
    
        & pick\_up\_the\_200ml\_beaker
        & $0.06$ & $0.36$ & $1.00$ \\
    
        & pick\_up\_the\_test\_tube
        & $0.00$ & $0.08$ & $0.06$ \\
    
        & take\_the\_glass\_rod\_from\_the\_white\_test\_tube\_rack
        & $0.02$ & $0.24$ & $0.22$ \\
    
        & pick\_up\_the\_funnel\_from\_the\_transparent\_funnel\_stand
        & $0.00$ & $0.84$ & $0.88$ \\
    
        & pick\_up\_the\_graduated\_cylinder
        & $0.00$ & $0.20$ & $0.04$ \\
    
        & pick\_up\_the\_weighing\_funnel
        & $0.00$ & $0.00$ & $0.04$ \\
    
        & remove\_the\_funnel\_from\_the\_round\_bottom\_flask
        & $0.00$ & $0.40$ & $0.52$ \\
    
        & \textbf{Average}
        & $0.05$ & $0.25$ & $\mathbf{0.46}$ \\
    
        \midrule
    
        \multirow{7}{*}{Place}
        & place\_the\_round\_bottom\_flask\_on\_the\_heater
        & $0.08$ & $0.00$ & $0.78$ \\
    
        & place\_the\_beaker\_back\_on\_the\_table
        & $0.00$ & $0.96$ & $0.96$ \\
    
        & put\_the\_100ml\_beaker\_on\_the\_electronic\_balance
        & $0.00$ & $0.56$ & $0.94$ \\
    
        & put\_the\_200ml\_beaker\_on\_the\_electronic\_balance
        & $0.00$ & $0.74$ & $0.90$ \\
    
        & put\_the\_graduated\_cylinder\_back\_on\_the\_table
        & $0.00$ & $0.68$ & $0.92$ \\
    
        & put\_the\_weighing\_funnel\_back\_on\_the\_table
        & $0.00$ & $0.88$ & $1.00$ \\
    
        & \textbf{Average}
        & $0.01$ & $0.64$ & $\mathbf{0.92}$ \\
    
        \midrule
    
        \multirow{5}{*}{Insert}
        & put\_the\_test\_tube\_back\_in\_the\_rack
        & $0.00$ & $0.08$ & $0.26$ \\
    
        & put\_the\_glass\_rod\_back\_on\_the\_white\_test\_tube\_rack
        & $0.00$ & $0.60$ & $0.52$ \\
    
        & place\_the\_funnel\_on\_the\_round\_bottom\_flask
        & $0.00$ & $0.02$ & $0.20$ \\
    
        & put\_the\_funnel\_back\_on\_the\_funnel\_stand
        & $0.00$ & $0.28$ & $0.22$ \\
    
        & \textbf{Average}
        & $0.00$ & $0.25$ & $\mathbf{0.30}$ \\
    
        \midrule
    
        \multirow{4}{*}{Pour}
        & pour\_the\_water\_from\_the\_handheld\_beaker\_into\_the\_one\_on\_the\_table
        & $0.00$ & $0.70$ & $0.94$ \\
    
        & pour\_the\_water\_from\_the\_graduated\_cylinder\_into\_the\_round\_bottom\_flask
        & $0.00$ & $0.00$ & $0.78$ \\
    
        & pour\_all\_the\_NaCl\_from\_the\_weighing\_funnel\_into\_the\_round\_bottom\_flask
        & $0.00$ & $0.00$ & $0.26$ \\
    
        & \textbf{Average}
        & $0.00$ & $0.23$ & $\mathbf{0.66}$ \\
    
        \midrule
    
        \multirow{2}{*}{Press}
        & turn\_on\_the\_heater
        & $0.00$ & $0.00$ & $0.12$ \\
    
        & \textbf{Average}
        & $0.00$ & $0.00$ & $\mathbf{0.12}$ \\
    
        \midrule
    
        \multirow{2}{*}{Shake}
        & shake\_the\_test\_tube
        & $0.00$ & $0.90$ & $0.88$ \\
    
        & \textbf{Average}
        & $0.00$ & $\mathbf{0.90}$ & $0.88$ \\
    
        \midrule
    
        \multirow{2}{*}{Stir}
        & stir\_in\_the\_beaker\_to\_the\_left\_of\_the\_white\_test\_tube\_rack\_with\_the\_glass\_rod
        & $0.00$ & $0.00$ & $0.46$ \\
    
        & \textbf{Average}
        & $0.00$ & $0.00$ & $\mathbf{0.46}$ \\
    
        \midrule
    
        \multicolumn{2}{l}{\textbf{Overall Average}}
        & $0.02$ & $0.34$ & $\mathbf{0.57}$ \\
    
        \bottomrule
    \end{tabular*}

\end{table*}

\begin{table*}[!t]
    \centering
    \caption{Performance of atomic skills in simulation. Each task evaluated over 50 trials for every model.}
    \label{tab:sim_atomic_skill_result}
    \small
    \setlength{\tabcolsep}{3pt}
    \renewcommand{\arraystretch}{1.15}

    \begin{tabular*}{\textwidth}{
        @{\extracolsep{\fill}}
        lccc
        @{}
    }
        \toprule
        \textbf{Task}
        & \textbf{DP}
        & \textbf{ACT}
        & ${\pi}_{0.5}$ \\
        \midrule

        pick\_up\_the\_round\_bottom\_flask
        & $0.00$ & $0.00$ & $0.56$ \\

        pick\_up\_the\_beaker
        & $0.34$ & $0.22$ & $0.64$ \\

        put\_the\_beaker\_on\_the\_electronic\_balance
        & $0.08$ & $0.32$ & $0.02$ \\

        pick\_up\_the\_graduated\_cylinder
        & $0.00$ & $0.04$ & $0.26$ \\

        pour\_the\_water\_from\_the\_graduated\_cylinder\_into\_the\_round\_bottom\_flask
        & $0.52$ & $0.74$ & $0.72$ \\

        put\_the\_graduated\_cylinder\_back\_on\_the\_table
        & $0.00$ & $0.08$ & $0.92$ \\

        \midrule

        \textbf{Overall Average}
        & $0.19$ & $0.26$ & $\mathbf{0.44}$ \\

        \bottomrule
    \end{tabular*}
\end{table*}

\section{Dataset Details}

\subsection{Laboratory Objects and Assets}

LabDex incorporates a diverse collection of commonly used chemistry laboratory objects and supporting equipment. The manipulated objects include beakers of different sizes, round-bottom flasks, Erlenmeyer flasks, graduated cylinders, test tubes, glass rods, standard funnels, and a weighing funnel. The supporting assets include funnel stands, test-tube racks, a round-bottom-flask holder, cork rings, an electronic balance, and a laboratory heater. Several task scenes also contain multiple instances of similar objects, such as two beakers, to better reflect realistic laboratory settings and increase the difficulty of target-object selection.

These objects cover a wide range of geometries and manipulation requirements, including cylindrical and elongated objects, narrow-necked containers, transparent glassware, and objects with limited effective grasping regions. Together, they support diverse laboratory operations such as grasping, placing, insertion, pouring, stirring, weighing, and object transfer. Corresponding assets are constructed in the real-world and simulation platforms to maintain consistent task definitions and hierarchical task structures across both settings.

\subsection{Atomic-Skill}
\paragraph{Task Definitions.}
LabDex defines Atomic Skills as fundamental operations with explicit objectives that can be independently evaluated and reused across different laboratory tasks.  The task name describes the operation to be performed and the corresponding object, such as picking up a 200 ml beaker, inserting a test tube into a test-tube rack, or pouring material from a graduated cylinder into a round-bottom flask. The initial state specifies the configurations of the robot and relevant laboratory objects at the beginning of the task, while the goal state describes the required object state or spatial relationship upon task completion. A task is considered successful only when the goal state satisfies the predefined tolerance and remains stable for the specified duration.

\paragraph{Atomic-Skill Categories.}
Atomic Skill Categories. We group all atomic skills into seven categories according to their operation types and provide a concise description of each task below.

\paragraph{Pick.}
Pick tasks require the robot to grasp a target object and remove it from its initial support or holder while maintaining a stable grasp.

\begin{itemize}
    \item pick up the round bottom flask: Grasp the round-bottom flask and lift it from its holder.
    
    \item pick up the beaker on the right: Grasp the beaker on the right and lift it from the table.
    
    \item pick up the 100 ml beaker: Grasp the 100 ml beaker and lift it from the table.
    
    \item pick up the beaker: Grasp the target beaker and lift it from the table.
    
    \item pick up the test tube: Grasp the test tube and remove it from the test-tube rack.
    
    \item take the glass rod from the white test tube rack: Grasp the glass rod and remove it from the white test-tube rack.
    
    \item pick up the funnel from the transparent funnel stand: Grasp the funnel and remove it from the transparent funnel stand.
    
    \item pick up the graduated cylinder: Grasp the graduated cylinder and lift it from the table.
    
    \item pick up the weighing funnel: Grasp the weighing funnel and lift it from the table.
    
    \item remove the funnel from the round bottom flask: Grasp the funnel and remove it from the round-bottom flask.
\end{itemize}

\paragraph{Place.}
Place tasks require the robot to move a grasped object to a designated target region or laboratory device and release it in a stable state.

\begin{itemize}
    \item place the round bottom flask on the heater: Place the round-bottom flask on the heater and release it stably.
    
    \item place the beaker back on the table: Return the grasped beaker to the table and release it stably.
    
    \item put the 100 ml beaker on the electronic balance: Place the 100 ml beaker on the electronic balance.
    
    \item put the beaker on the electronic balance: Place the target beaker on the electronic balance.
    
    \item put the graduated cylinder back on the table: Return the graduated cylinder to the table and release it stably.
    
    \item put the weighing funnel back on the table: Return the weighing funnel to the table and release it stably.
\end{itemize}

\paragraph{Insert.}
Insert tasks require the robot to align a manipulated object with a target opening, rack, or holder and insert it into the designated position.

\begin{itemize}
    \item put the test tube back in the rack: Align the test tube with the target slot and insert it into the test-tube rack.
    
    \item put the glass rod back on the white test tube rack: Return the glass rod to its designated position on the white test-tube rack.
    
    \item place the funnel on the round bottom flask: Align the funnel with the opening of the round-bottom flask and insert it into the flask.
    
    \item put the funnel back on the funnel stand: Align the funnel with the funnel stand and return it to the designated position.
\end{itemize}

\paragraph{Pour.}
Pour tasks require the robot to maintain a stable grasp on the source container while adjusting its orientation to transfer the contained material into a target container.

\begin{itemize}
    \item pour the water from the handheld beaker into the one on the table: Pour water from the grasped beaker into the beaker on the table.
    
    \item pour the water from the graduated cylinder into the round bottom flask: Pour water from the graduated cylinder into the round-bottom flask.
    
    \item pour all the NaCl from the weighing funnel into the round bottom flask: Transfer all NaCl from the weighing funnel into the round-bottom flask.
\end{itemize}

\paragraph{Stir.}
Stir tasks require the robot to manipulate a glass rod continuously inside a container while maintaining stable control of the tool.

\begin{itemize}
    \item stir in the beaker to the left of the white test tube rack with the glass rod: Use the glass rod to stir the contents of the beaker located to the left of the white test-tube rack.
\end{itemize}

\paragraph{Press.}
Press tasks require the robot to accurately contact and activate a target control on a laboratory device.

\begin{itemize}
    \item turn on the heater: Press the control button to turn on the heater.
\end{itemize}

\paragraph{Shake.}
Shake tasks require the robot to maintain a stable grasp while repeatedly moving the manipulated object.

\begin{itemize}
    \item shake the test tube: Maintain a stable grasp on the test tube and repeatedly move it to shake its contents.
\end{itemize}

\begin{figure*}[t]
    \centering

    \begin{subfigure}[t]{0.32\textwidth}
        \centering
        \includegraphics[width=\linewidth]{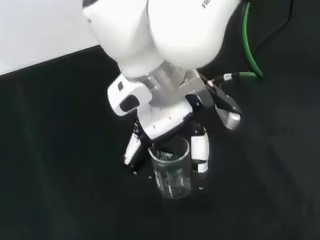}
        \caption{pick up the 200ml beaker}
        \label{fig:subfig_a}
    \end{subfigure}
    \hfill
    \begin{subfigure}[t]{0.32\textwidth}
        \centering
        \includegraphics[width=\linewidth]{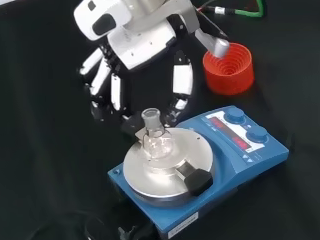}
        \caption{place the round bottom flask on the heater}
        \label{fig:subfig_b}
    \end{subfigure}
    \hfill
    \begin{subfigure}[t]{0.32\textwidth}
        \centering
        \includegraphics[width=\linewidth]{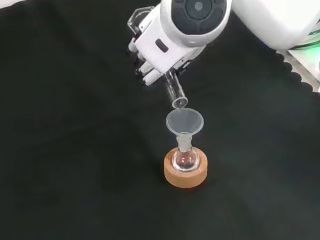}
        \caption{pour the water from the graduated cylinder into the round bottom flask}
        \label{fig:subfig_c}
    \end{subfigure}

    \vspace{3mm}

    \begin{subfigure}[t]{0.32\textwidth}
        \centering
        \includegraphics[width=\linewidth]{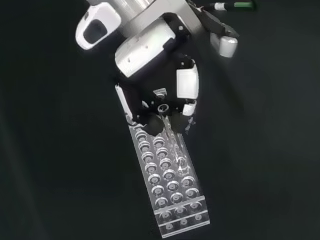}
        \caption{put the test tube back in the rack}
        \label{fig:subfig_d}
    \end{subfigure}
    \hfill
    \begin{subfigure}[t]{0.32\textwidth}
        \centering
        \includegraphics[width=\linewidth]{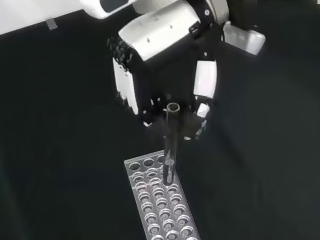}
        \caption{shake the test tube}
        \label{fig:subfig_e}
    \end{subfigure}
    \hfill
    \begin{subfigure}[t]{0.32\textwidth}
        \centering
        \includegraphics[width=\linewidth]{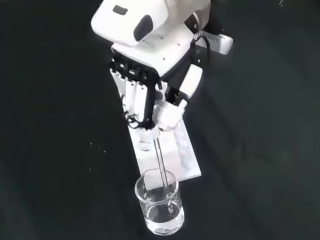}
        \caption{stir in the beaker to the left of the white test tube rack with the glass rod}
        \label{fig:subfig_f}
    \end{subfigure}

    \caption{Examples of selected atomic skills.}
    \label{fig:atomic_task_visualizations}
\end{figure*}

Figure~\ref{fig:atomic_task_visualizations} presents representative examples of Atomic Skills from different operation categories.

\subsection{Compositional-Skill}

Compositional Skills consist of multiple Atomic Skills executed sequentially to complete a laboratory operation with a specific functional objective. The complete set of compositional tasks is described below. For each task, we list the constituent Atomic Skills in their execution order.

\begin{itemize}
    \item place the round-bottom flask on the heater and turn it on: The robot picks up the round-bottom flask, places it on the heater, and turns on the heater.

    \item pour the water from the right beaker into the left one: The robot picks up the beaker on the right, pours the water into the beaker on the left, and returns the grasped beaker to the table.

    \item put the 100 ml beaker on the electronic balance: The robot picks up the 100 ml beaker and places it on the electronic balance.

    \item put the 200 ml beaker on the electronic balance: The robot picks up the 200 ml beaker and places it on the electronic balance.

    \item shake the test tube and put it back in the rack: The robot picks up the test tube, shakes it, and returns it to the test-tube rack.

    \item stir the solution in the beaker with a glass rod: The robot removes the glass rod from the white test-tube rack, stirs the solution in the beaker, and returns the glass rod to the rack.

    \item pour the water from the graduated cylinder into the round-bottom flask: The robot picks up the graduated cylinder, pours the water into the round-bottom flask, and returns the graduated cylinder to the table.

    \item place the funnel on the round-bottom flask: The robot removes the funnel from the transparent funnel stand and inserts it into the round-bottom flask.

    \item pour the NaCl from the weighing funnel into the round-bottom flask: The robot picks up the weighing funnel, pours all the NaCl into the round-bottom flask, and returns the weighing funnel to the table.

    \item put the funnel back on the funnel stand: The robot removes the funnel from the round-bottom flask and returns it to the funnel stand.
\end{itemize}

\subsection{Long-Horizon Laboratory Workflow}

The Long-Horizon Laboratory Workflow consists of multiple Compositional Skills executed sequentially to complete a laboratory procedure. Compared with individual Compositional Skills, this workflow involves more objects, a longer operation sequence, and stronger dependencies between successive stages. The robot must maintain consistent object states throughout the workflow, as failures in an early stage may prevent subsequent operations from being executed.

\paragraph{Add the NaCl from the weighing funnel to the round-bottom flask.}
The robot first removes the funnel from the transparent funnel stand and places it on the round-bottom flask. It then picks up the weighing funnel, pours all the NaCl into the round-bottom flask through the funnel, and returns the weighing funnel to the table. Finally, the robot removes the funnel from the round-bottom flask and returns it to the funnel stand.

The workflow consists of the following Atomic Skills:

\begin{itemize}
    \item pick up the funnel from the transparent funnel stand;
    \item place the funnel on the round-bottom flask;
    \item pick up the weighing funnel;
    \item pour all the NaCl from the weighing funnel into the round-bottom flask;
    \item put the weighing funnel back on the table;
    \item remove the funnel from the round-bottom flask;
    \item put the funnel back on the funnel stand.
\end{itemize}

\end{document}